\documentclass[10pt,twocolumn,letterpaper]{article}

\usepackage[pagenumbers]{cvpr} 

\usepackage[dvipsnames]{xcolor}

\newcommand{\figref}[1]{Fig.~\ref{#1}}%
\newcommand{\tabref}[1]{Tab.~\ref{#1}}%

\renewcommand{\eqref}[1]{Eqn.~(\ref{#1})}

\usepackage{times}
\usepackage{graphicx}
\usepackage{overpic}
\usepackage{amsmath}
\usepackage{amssymb}
\usepackage{svg}
\usepackage{amsmath}
\usepackage{algorithm}
\usepackage{listings}

\usepackage{etoolbox}
\makeatletter
\AfterEndEnvironment{algorithm}{\let\@algcomment\relax}
\AtEndEnvironment{algorithm}{\kern2pt\hrule\relax\vskip3pt\@algcomment}
\let\@algcomment\relax
\newcommand\algcomment[1]{\def\@algcomment{\footnotesize#1}}
\renewcommand\fs@ruled{\def\@fs@cfont{\bfseries}\let\@fs@capt\floatc@ruled
  \def\@fs@pre{\hrule height.8pt depth0pt \kern2pt}%
  \def\@fs@post{}%
  \def\@fs@mid{\kern2pt\hrule\kern2pt}%
  \let\@fs@iftopcapt\iftrue}
\makeatother

\newcommand\blfootnote[1]{\begingroup\renewcommand\thefootnote{}\footnote{#1}\addtocounter{footnote}{-1}\endgroup}

\usepackage{makecell}

\definecolor{cvprblue}{rgb}{0.21,0.49,0.74}
\usepackage[pagebackref,breaklinks,colorlinks,citecolor=cvprblue]{hyperref}

\usepackage{overpic}
\usepackage{booktabs}
\usepackage{multirow}
\usepackage{pifont}
\usepackage{enumitem}
\usepackage{bm}
\usepackage{diagbox}

\graphicspath{{./figs/}}
\DeclareGraphicsExtensions{.pdf,.jpg,.png}

\newcommand{\ourMthd}{K-LoRA}
\newcommand{\myPara}[1]{\vspace{5pt}\noindent\textbf{#1}}

\def\paperID{13290} 
\def\confName{CVPR}
\def\confYear{2025}
\definecolor{lightgreen}{RGB}{64,128,128}

\title{\ourMthd{}: Unlocking Training-Free Fusion of Any Subject and Style LoRAs}

\author{Ziheng Ouyang \quad Zhen Li\textsuperscript{$\dagger$} \quad Qibin Hou\\
VCIP, School of Computer Science, Nankai University\\
{\tt\small \{zihengouyang666, zhenli1031\}@gmail.com} \\ \\
{ Project page: \url{https://k-lora.github.io/K-LoRA.io/}}
}

\begin{document}

\twocolumn[{%
\renewcommand\twocolumn[1][]{#1}%
\maketitle
\begin{center}
   \captionsetup{type=figure}
    \vspace{-0.8cm}    
\includegraphics[width=0.9\linewidth]{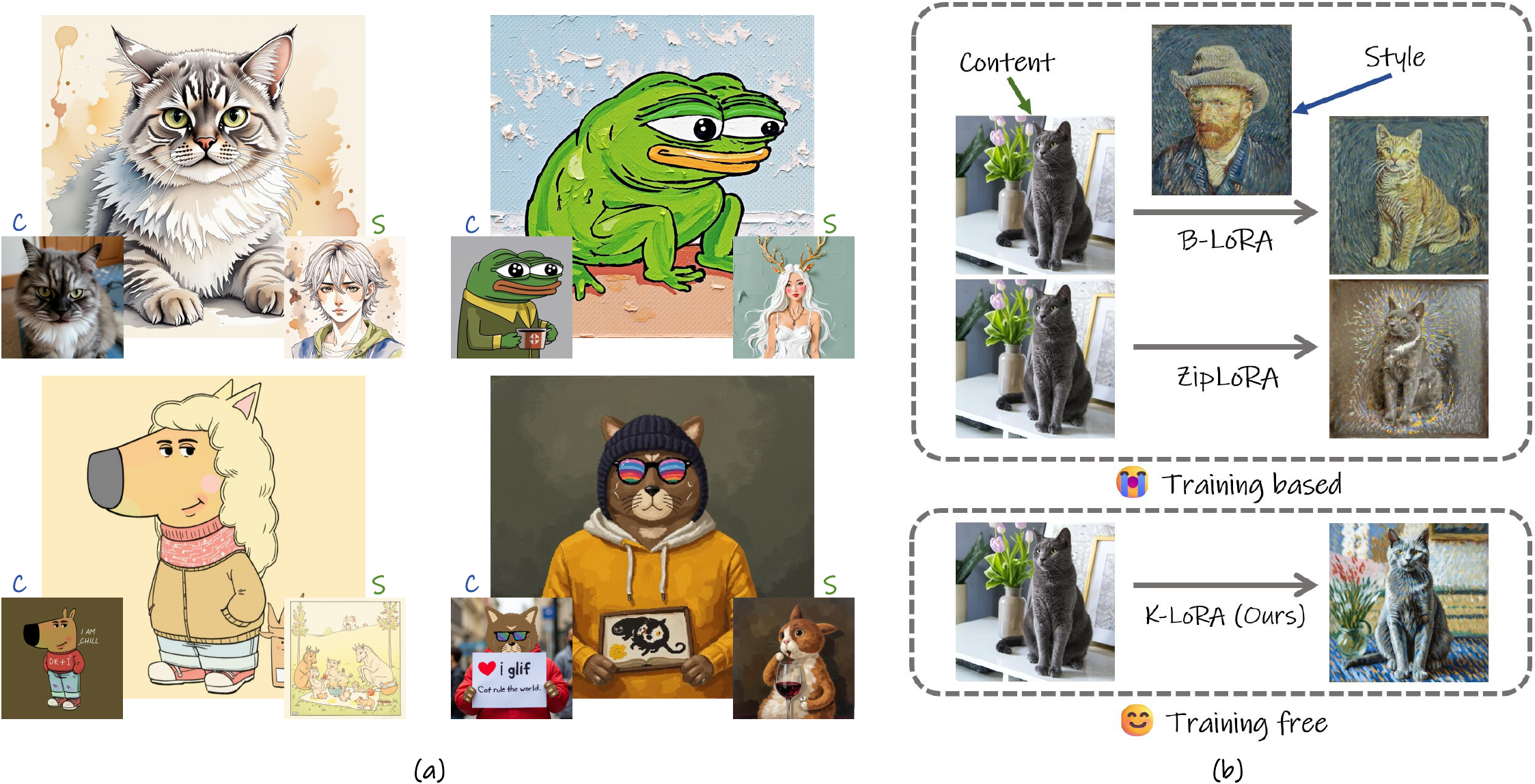}    
    \vspace{-0.2cm}
    \caption{\textbf{Visual illustrations.} (a) demonstrates the superior generative performance of our proposed \ourMthd{} using FLUX~\cite{flux}, where the object reference is presented on the left, the style reference on the right, and the generated image is shown in the center. In contrast, (b) compares our method with existing state-of-the-art methods, B-LoRA~\cite{B-LoRA} and ZipLoRA~\cite{ZipLoRA}, which tend to lose style or content information due to alterations in the original weight matrix or underutilization of the network structure. Our approach enhances the information captured by each LoRA matrix, thereby achieving superior fusion effects \textit{without requiring additional training.}
    }
    \label{fig:teaser} 
\end{center}   
}]

\begin{abstract}
Recent studies have explored the combination of different LoRAs to jointly generate learned style and content.
However, existing methods either fail to effectively preserve both the original subject and style simultaneously or require additional training.
In this paper, we argue that the intrinsic properties of LoRA can effectively guide diffusion models in merging learned subject and style. 
Based on this insight, we propose \ourMthd{}, a simple yet effective training-free LoRA fusion approach. 
In each attention layer, \ourMthd{} compares the Top-K elements in each LoRA to be fused, determining which LoRA to select for optimal fusion.
This selection mechanism ensures that the most representative features of both subject and style are retained during the fusion process, effectively balancing their contributions. 
Experiments demonstrate that \ourMthd{} can effectively integrates the subject and style information learned by the original LoRAs, outperforming state-of-the-art training-based approaches in both qualitative and quantitative results.

\end{abstract}

\blfootnote{$^{\dagger}$ Corresponding authors.}

\begin{figure*}[]
    \centering
    \setlength{\abovecaptionskip}{0pt}
    \includegraphics[width=\linewidth]{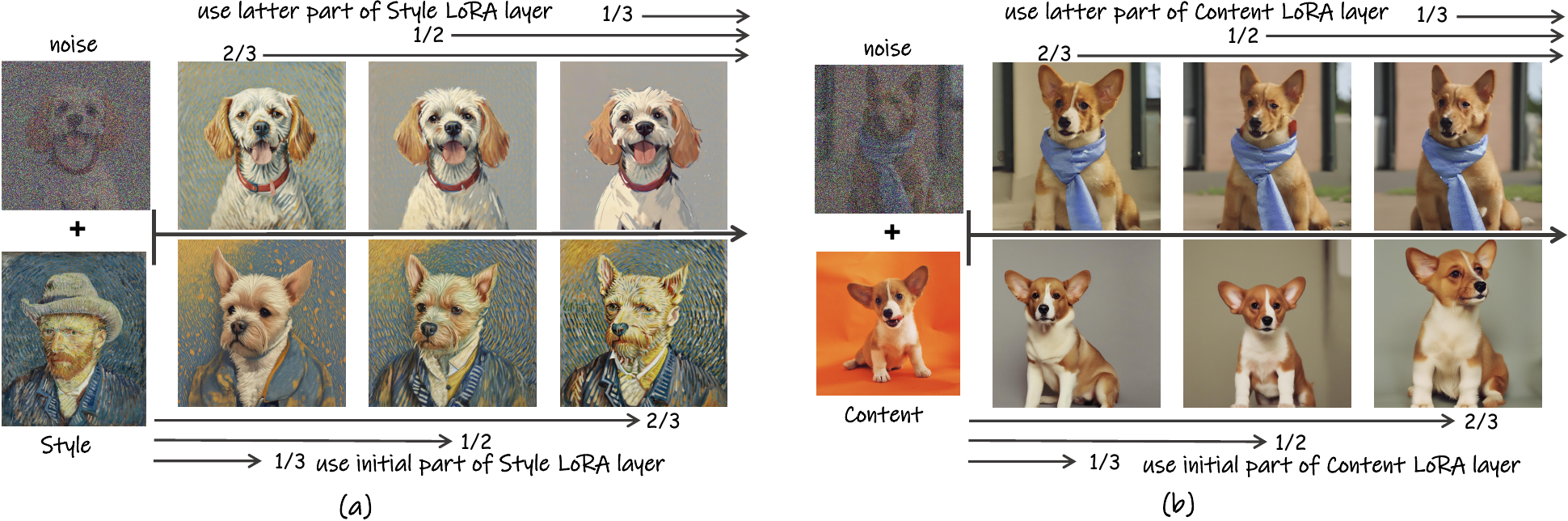}
    \caption{\textbf{Visual results of findings.} (a) Results fine-tuned using only content LoRA. (b) Results only using style LoRA. In these experiments, we test the differences between adding the LoRA layers in the initial and latter timesteps.
    }
    \label{fig:insight_evidence}
\end{figure*}

\section{Introduction}
\label{sec:intro}

Personalization and stylization are two well-established tasks in computer vision and have been active research fields for many years \cite{gatys2016image,chen2017stylebank,huang2017arbitrary,photomaker,ruiz2023dreambooth,ye2023ip,sohn2023styledrop,xing2024csgo,zhou2024magictailor,deng2020arbitrary}. 
The primary challenge in these tasks is preserving distinct content or modifying the style of an image, typically guided by textual or visual inputs. 
In this context, ``content" refers to the objects and structures within the image, while ``style" encompasses visual attributes such as color, texture, and patterns.
Manipulating image style is particularly challenging due to the subjective nature of style definitions and the strong interdependence between style and content, which complicates effective decoupling of these elements.

Recent techniques, such as LoRA~\cite{LoRA}, have gained more and more attention for their ability to achieve efficient fine-tuning in image synthesis. 
Although styles and objects are trained separately, LoRA provides an effective solution to the problems of decoupling style and content in image generation tasks, which excels in controlling style transfer by training style features independently from the content features.
With the growing popularity of personalized applications utilizing LoRA, numerous efforts have been put to fuse objects and styles by merging LoRA weights~\cite{Mergingloras}.
These approaches aim to allow users to adjust each LoRA’s contribution ratio through variable coefficients. There are also methods, such as ZipLoRA~\cite{ZipLoRA}, attempting to train a fusion ratio vector to balance different LoRAs.
More recently, some approaches propose the periodic integration of LoRA into models~\cite{Multi-LoRAComposition}. 
Additionally, the B-LoRA~\cite{B-LoRA} technique fine-tunes only two attention modules to facilitate style transfer.

In our experiments with these methods, we identify two key issues, as shown in \figref{fig:teaser}(b). First, \emph{style details often lose in the generated images, and the object characteristics are inconsistently maintained.}
Second, \emph{manual tuning of certain hyperparameters and seeds is required, or additional training is necessary.}
For the first issue, we conduct extensive experiments and observe that merging the attention layers of two LoRAs at the element level could lead to a smoothing of style details and textures, or even the loss of object characteristics. 
Given that element-level merging may lead to suboptimal results, we conduct experiments by selectively removing certain elements to keep good performance. 
For the second issue, inspired by the core ideas in recent studies~\cite{xu2024freetuner,voynov2023p+,patashnik2023localizing}, we incorporate the attention layers of LoRA into the model according to diffusion time steps to assess their effects on performance. 
Through this approach, we derive key conclusions. (i) Only a restricted number of diffusion prediction steps are sufficient to retain the original effect as illustrated in~\figref{fig:insight_evidence}. (ii) When applying LoRA, the initial diffusion steps are responsible for reconstructing the object and capturing larger texture details, while the subsequent steps focus on enhancing and refining the finer details of the object and the texture in style.

Based on these findings, we propose \ourMthd{}, which simultaneously addresses both issues identified in our experiments, as illustrated in \figref{fig:teaser}(a), leveraging our first insight by incorporating a Top-K selection process within each forward pass of the attention layers to identify the most suitable attention components at each position. 
Additionally, we apply a scaling factor during the selection process, utilizing our second insight to emphasize the distinct roles that style and content play throughout the diffusion process.

Our method can effectively resolve the aforementioned issues, ensuring that the merged LoRA captures both subject and stylistic features when faced with challenging content and style combinations. This results in stable generative outputs and significantly enhances the performance of merged LoRAs. Furthermore, our approach is user-friendly, as it requires no additional training. We summarize our contributions as follows:
\begin{itemize}
    \item We propose \ourMthd{}, a simple yet effective optimization technique that seamlessly merges content and style LoRAs, enabling the generation of any desired style for any theme while preserving intricate details.
    \item Our method is user-friendly, eliminating the need for retraining and directly applicable to existing LoRA weights. It demonstrates superior performance across diverse image stylization tasks, surpassing existing methods.
\end{itemize}

\section{Related Work}
\label{sec:relatedwork}

\myPara{Diffusion models for customization.} In the realm of diffusion models~\cite{rombach2022high} for customized tasks, customization refers to the process by which the model learns to interpret new definitions provided by the user.
Techniques such as Textual Inversion~\cite{zhang2023inversion,alaluf2023neural,voynov2023p+}, DreamBooth~\cite{ruiz2023dreambooth}, and Custom Diffusion~\cite{kumari2023multi} enable the model to capture target concepts with only a limited number of images through token-based optimization. 
Specifically, Textual Inversion fine-tunes embeddings to reconstruct the target, DreamBooth uses less common class-specific terms to expand object categories, and Custom Diffusion focuses on fine-tuning the cross-attention layers within the diffusion model to learn new concepts. 
Additionally, there are methods that do not require training 
when inferring~\cite{xie2023smartbrush,xiao2024fastcomposer,avrahami2023break,shi2024instantbooth}, but their approaches to utilize pre-trained modules may perform suboptimally for certain specialized tasks.
LoRA~\cite{LoRA} and its variants~\cite{hayou2024lora+,zhang2023lora,zhou2024lora,zhou2024lora,kopiczko2023vera,zi2023delta,ren2024melora} are well-known for their ability to fine-tune large models and deliver high-quality results, making them an good choice for practitioners.

\myPara{LoRA combination in image generation.} In the field of image generation, research on LoRA combinations has primarily been advanced in two directions, including the integration of multiple objects and the fusion of contents with styles.
For object integration, studies have mainly focused on enabling models to integrate diverse object concepts encapsulated within multiple LoRAs~\cite{gu2024mix,LoRA-Composer,jiang2024mc,dong2024continually,liu2023cones}. 
By fine-tuning the subject LoRAs, these models can assimilate various new concepts and manage object layouts through masking techniques. 
Regarding content-style fusion, several works, such as MergingLoRA~\cite{Mergingloras}, Mixture-of-Subspaces~\cite{wu2024mixture}, and ZipLoRA~\cite{ZipLoRA}, have proposed approaches involving hyperparameter tuning and learning fusion matrices to merge pre-trained LoRA weight layers.
However, these methods may face challenges, such as concept dilution, blurring of fine details, and specific training requirements. 
Recently, B-LoRA~\cite{B-LoRA} has identified distinct roles for attention modules in the generative process, thereby achieving object-style decoupling within LoRA by training only two core attention modules. 
Additionally, LoRA Composition~\cite{Multi-LoRAComposition} uses a cyclic update of the model's LoRA modules to allow multiple LoRAs to collaboratively guide the model, allowing a variety of cross-concept fusion. 
Despite these advancements, existing methods continue to face challenges, including insufficient control precision, loss of object style, and high training requirements.

\section{Method}

\subsection{Preliminaries}

LoRA is an effective method initially designed to adapt large-scale language models. 
The core premise of LoRA is that, when fine-tuning a large model and comparing it with a baseline model, the parameter update matrix $\Delta W \in \mathbb{R}^{m \times n}$ is typically found to contain small or near-zero elements, exhibiting a low-rank structure. 
This property allows $\Delta W$ to be factorized into two low-rank matrices, $B \in \mathbb{R}^{m \times r}$ and $A \in \mathbb{R}^{r \times n}$, where $r$ represents the intrinsic rank of $\Delta W$, and it is assumed that $r \ll \min(m, n)$.
This characteristic enables us to freeze the base weight $W_0$ and train only the matrices $B$ and $A$ to replace $\Delta W$, thereby achieving an efficient parameterization in the form $\Delta W = BA$.
Finally, $\Delta W$ is added to the base weight in the original model to perform fine-tuning. The updated weights can be expressed as $W_0 + \Delta W$.

In our work, we adopt the same notation as used in ZipLoRA ~\cite{ZipLoRA}.
Let $D$ be a base diffusion model, and $W_0$ denote the pre-trained weights that need to be updated with LoRA layer.
The base model $D$ can be adapted to a specific concept simply by adding an additional trained LoRA weight set $ \Delta W_x $ to the model weights, resulting in $D' = W_0 + \Delta W_x$.
Given two independently trained LoRA weight sets, $\Delta W_c$ and 
$\Delta W_s$, associated with the base model $D$, our objective is to fully leverage the weights of both LoRA sets and enable their effective fusion. To achieve this, we propose a method, called \ourMthd{}, to seamlessly combine the two LoRA weight sets, expressed as 
\[
\Delta W_x = K(\Delta W_c, \Delta W_s),
\]
where $K$ denotes our method, which can efficiently integrate the contributions of the content LoRA and style LoRA.

In what follows, we will explain the proposed approach in detail.
Our approach is based on two findings. (i) In the diffusion steps, applying LoRA to only a subset of layers per step can achieve comparable effects comparing to applying LoRA to all layers; (ii) Using the subject LoRA in earlier diffusion steps tends to generate better subject information, while using the style LoRA in later steps is more effective for generating style and details without affecting the construction of the content.

\subsection{\ourMthd{}}

\begin{figure}
    \centering
    \includegraphics[width=\linewidth]{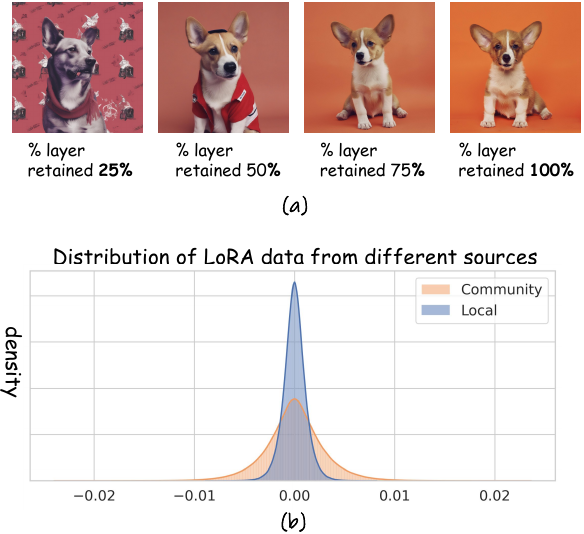}
    \caption{\textbf{Experimental visualization results.} (a) Generated images by randomly loading a portion of the LoRA attention layers according to a certain ratio. (b) Visualization of LoRA data distribution from different sources: one trained locally and the other one downloaded from a community repository.}
    \label{fig:data_distribution}
\end{figure}

\begin{figure*}[t]
    \centering   \includegraphics[width=0.9\textwidth]{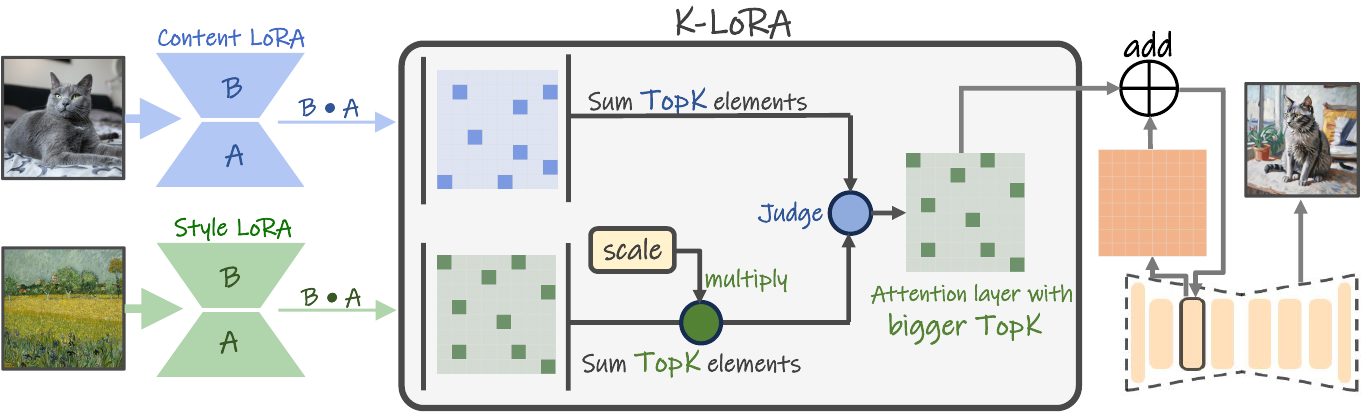}
    \caption{\textbf{Overview of the proposed \ourMthd{}}. We propose \ourMthd{}, which utilizes the Top-K function to select the important LoRA weights in each forward layer based on the sum of matrix elements. This enables us to preserve both stylistic details and object features.} 
    \label{fig:net}
\end{figure*}

It has been pointed out in~\cite{ZipLoRA} that using a smaller set of key elements when finetuning with LoRA can achieve the same generative results as the original approach. However, the authors did not provide relevant experiments to explain this in the field of image generation.
We first attempt to leverage this method by assigning zeros to the elements whose values are small following a similar approach to that of Magmax~\cite{marczak2024magmax}. 
We found that the results obtained by modifying the elements of the matrix in this way are similar to the ones produced by~\cite{ZipLoRA, wu2024mixture} because the model does not correctly interpret the concepts it has previously learned, resulting in a suboptimal quality of image generation.

Given the complexities and limitations involved in directly modifying attention elements, a question rises: Can we exploit the sparse characteristics of the LoRA matrix during the denoising process?
The aim is to find an alternative method that can identify a good weight selection method and precise LoRA positioning for each step or layer without modifying the original LoRA weights.
Based on Multi-LoRA Composition~\cite{Multi-LoRAComposition},
we randomly apply the content LoRA attention layer to the diffusion steps, affecting the object using $x\%$ of the attention layers to observe the generated outcomes. 
As shown in~\figref{fig:data_distribution}(a),
we found that when $x > 50$, the results are virtually indistinguishable from those of the original model.
However, when $x < 25$, the ability of the model to maintain the original personalized concepts significantly diminished. 

Inspired by recent studies~\cite{xu2024freetuner, voynov2023p+, patashnik2023localizing}, we further extend the aforementioned experiments in~\figref{fig:insight_evidence} and found that applying the style LoRA in earlier timesteps has a significant impact on the reconstruction of the original object, whereas applying it in later timesteps preserves the style information without affecting the original object.
Additionally, we observe that for content LoRA, applying it in earlier timesteps yields significantly better results than applying it in later timesteps.

The above analysis motivates us to achieve the merging of generated objects and styles by adaptively selecting the LoRA module for each attention layer. 
According to finding (i), the selection strategy should preserve the overall object and style information. Furthermore, according to finding (ii), the generation process should be achieved by arranging the object and style components appropriately. That is in the early diffusion steps, the model should focus more on object reconstruction while introducing style textures, and in later steps, it is better to refine the style with subtle object details. 
Therefore, we present \ourMthd{}, as shown in \figref{fig:net}, which can adaptively select the appropriate LoRA layer for merging learned subject and style.

First, we take the absolute value of each element in the LoRA Layer to determine whether a particular value plays a significant role in the generation process,
\begin{align}
\Delta W_c' = |\Delta W_c|,\\
\Delta W_s' = |\Delta W_s|,
\end{align}
where $W_c$ and $W_s$ denote the content and style LoRA weights, respectively.
Because a small subset of dominant elements can achieve the original generation effect while the data distribution (see \figref{fig:data_distribution}(b)) shows that smaller elements occupy a large proportion of the positions, which will influence the selection of the important elements, we use a smaller number of the largest elements to represent the importance of each layer.

Specifically, we select the top $K$ elements with the highest values from $\Delta W_c'$ and $\Delta W_s'$, respectively. By accumulating the Top-K elements, we assess the importance of the two matrices at a given attention layer: 
\begin{align}
S_c &= \sum_{i \in \text{Top-K}(\Delta W_c')} \Delta W_{c,i}', \\
S_s &= \sum_{j \in \text{Top-K}(\Delta W_s')}\Delta W_{s,j}',
\end{align}
where $\text{Top-K}$ returns the indices of the largest $K$ values.
For the selection of $K$, we note that the rank number in the LoRA training process reflects, to some extent, the amount of information contained within the matrix. Thus, our choice of $K$ is aligned with the rank of each LoRA:
\begin{equation}
    K = {r_{c} \cdot r_{s}},
\end{equation}
where $r_{c}$ and $r_{s}$ represent the ranks of the content and style LoRA layers, respectively. This formulation allows us to determine the appropriate weights within an attention layer by comparing the two sums
\begin{equation}
C(S_c, S_s) =
    \begin{cases}
        \Delta W_c, & \text{if } S_c \geq S_s \\
        \Delta W_s. & \text{otherwise}
    \end{cases}
\end{equation}

\begin{figure}[t]
    \centering
    \includegraphics[width=\linewidth]{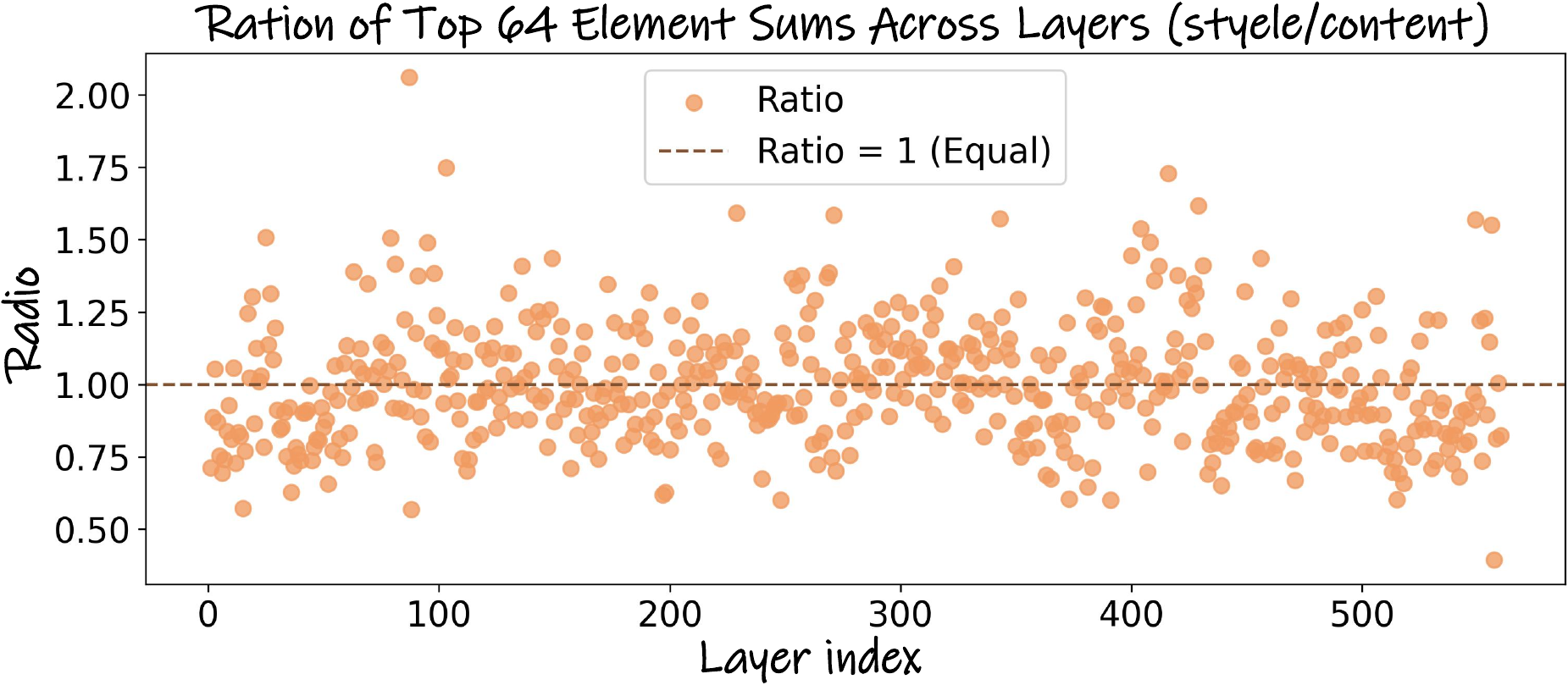}
    \caption{\textbf{Radio visualization.} This image reflects the ratio results after summing the Top-K elements, where the ratio differences at each corresponding position are quite significant.
    }
    \label{fig:dis}
\end{figure}

To more effectively leverage finding (ii) and allow both object and style to play their respective roles at different stages while ensuring a smooth transition from object-focused to style-focused representation, we introduce a scaling factor $S$.
This factor $S$ is directly applied to the Top-K selection process, enhancing object content in the early stages of generation and gradually emphasizing style in the later stages
\begin{equation}
    S = \alpha  \cdot \frac{{t}_{{now}}}{{t}_{{all}}} + \beta,
    \label{hypepara}
\end{equation}
where ${t}_{{now}}$ denotes the current step in the backward denoising process, ${t}_{{all}}$ is the total step number, and $\alpha, \beta$ are hyperparameters.

To avoid excessive weight disparities when using community LoRA models from different sources, which may make Top-K selection ineffective for attention allocation, we introduce a new factor $\gamma$ to balance the two weights
\begin{equation}
    S' = \gamma \cdot S.
\end{equation}
Initially, we compute the sum of the absolute values of the elements within each layer $l$, and then accumulate these sums layer by layer to calculate $\gamma$
\begin{equation}
\gamma = \frac
{\sum_{l} \sum_{i} \Delta W_{c_{l,i}}' }
{\sum_{l} \sum_{j} \Delta W_{s_{l,j}}' }.
\end{equation}

The introduction of $\gamma$ addresses the significant numerical discrepancy between the elements in the two LoRA components, as shown in \figref{fig:data_distribution}(b). This adjustment highlights the useful components within the LoRA layers. 
With $\gamma$, the proportional relationship between the content and style LoRA weights in each layer is shown in \figref{fig:dis}. It can be observed that, in each forward layer where LoRA is applied, there is a significant difference in the proportions of the dominant components' sums. This highlights the significance of the distinct LoRA weights within each layer, providing a solid basis for selection.

We then apply $S'$ to the style LoRA and update $S_s$
\begin{equation}
S_s' = S_s \cdot S'.
\end{equation}
By introducing $S'$, we can strengthen the influence of content during the earlier time steps, while amplifying the dominance of style in the later steps. 
This adjustment can effectively take advantage of finding (ii), optimizing the selection of both object and style to maximize their contributions in the image generation process.
The final LoRA weights can be attained by computing $C(S_c, S_s')$.
To clarify, we present the pseudo code in Algorithm~\ref{alg:code}.

\begin{figure*}[t]
    \centering
    \includegraphics[width=0.9\linewidth]{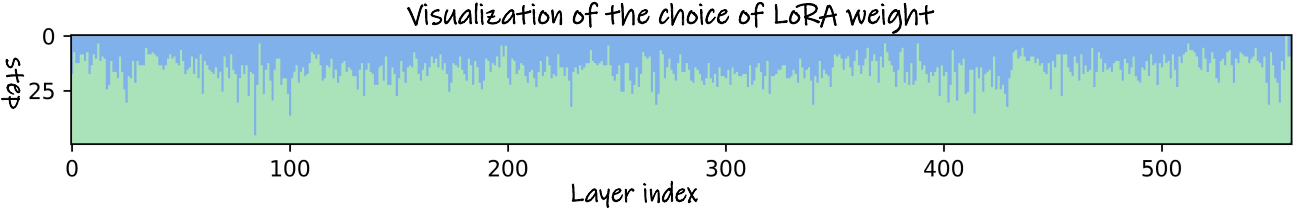}
    \caption{\textbf{LoRA selection during the generation process.} This figure illustrates the selection within each forward layer. The vertical axis represents the total 50 diffusion steps, while the horizontal axis indicates the number of LoRA layers. The color at each position denotes the selected layer. Blue bars correspond to objects, and green bars correspond to styles.
    }
    \label{fig:choice}
\end{figure*}

\begin{figure*}[h]
    \centering
    \includegraphics[width=\textwidth]{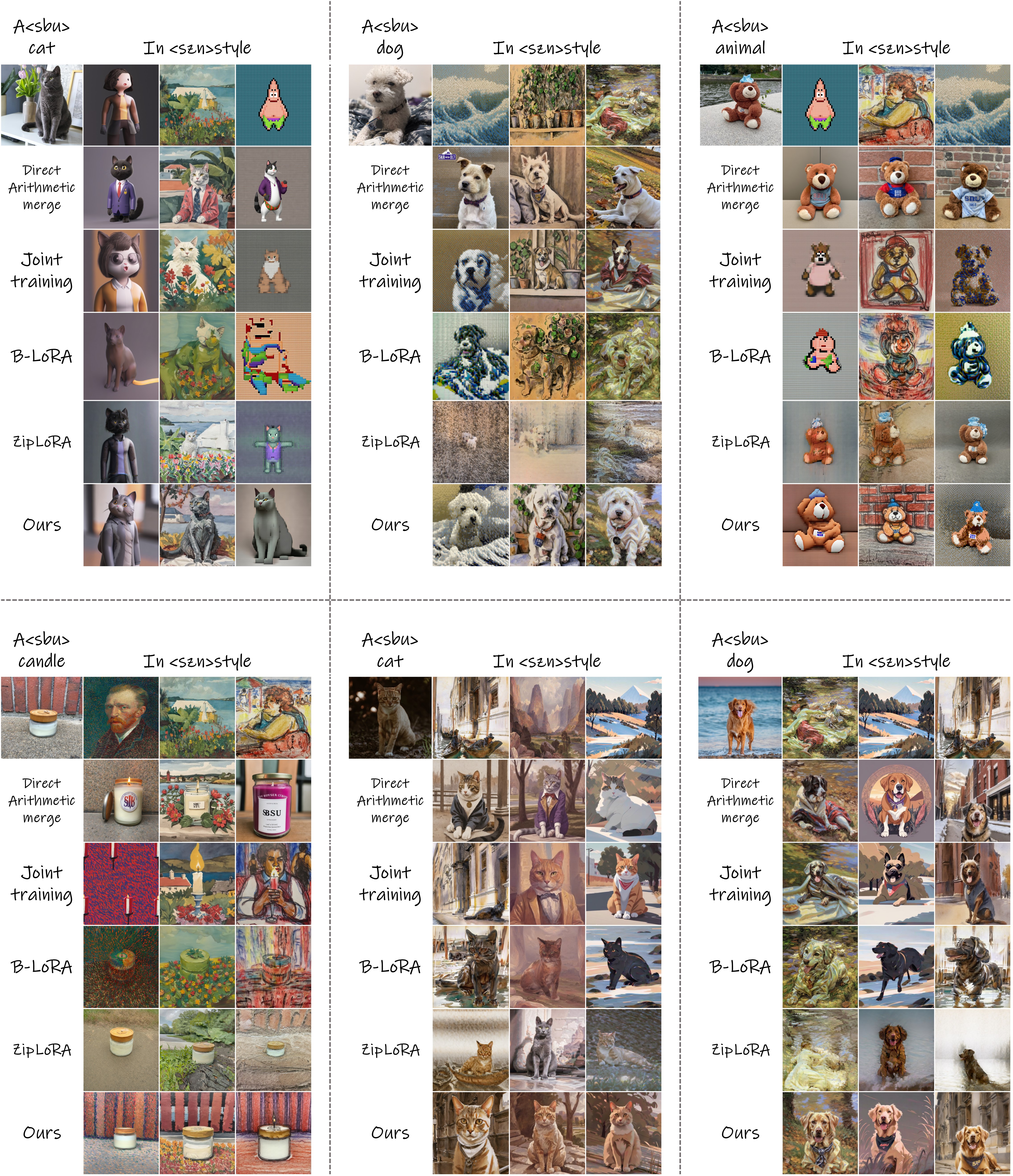}
    \caption{\textbf{Qualitative comparisons.} We present images generated by \ourMthd{} and the compared methods. \ourMthd{} generally achieves a seamless integration of objects and styles, effectively preserving fidelity and preventing distortion.
    }
    \label{fig:compare}
\end{figure*}

\begin{algorithm}[t]
\caption{Pseudocode in a PyTorch-like style.}
\label{alg:code}
\algcomment{\fontsize{7.2pt}{0em}\selectfont \texttt{fl}: flatten;}
\definecolor{codeblue}{rgb}{0.25,0.5,0.5}
\lstset{
  backgroundcolor=\color{white},
  basicstyle=\fontsize{7.2pt}{7.2pt}\ttfamily\selectfont,
  columns=fullflexible,
  breaklines=true,
  captionpos=b,
  commentstyle=\fontsize{7.2pt}{7.2pt}\color{codeblue},
  keywordstyle=\fontsize{7.2pt}{7.2pt},
}

\begin{lstlisting}[language=python]
# timestep: current timestep
# content_lora_weight, style_lora_weight: input weights
# alpha, beta, gamma: scaling factors
# all_timesteps: total timesteps

# Set k based on rank
k = rank * rank

# Sum of TopK content values
abs_content_matrix = abs(content_lora_weight)
topk_content_values = topk(abs_content_matrix.fl(), k)
sum_topk_content = sum(topk_content_values)

# Sum of TopK style values
abs_style_matrix = abs(style_lora_weight)
topk_style_values = topk(abs_style_matrix.fl(), k)
sum_topk_style = sum(topk_style_values)

# Compute and apply scaling factor
scale = alpha + beta * timestep / all_timesteps
scale = scale * gamma
sum_topk_style *= scale

# Compare and return the result
if sum_topk_content >= sum_topk_style:
    return content_lora_weight
else:
    return style_lora_weight
\end{lstlisting}
\end{algorithm}

To better explain the weight selection process, we show the selection proportions in~\figref{fig:choice}, where the object and style seamlessly interpenetrate and blend with each other.
The first portion primarily focuses on the object, with a small amount of style incorporated, while the latter portion predominantly emphasizes style, retaining a subtle presence of the object which further substantiates our key findings.

\section{Experiments}

\subsection{Experiment Setup}

\myPara{Datasets.}
Following the convention of ZipLoRA~\cite{ZipLoRA}, for the LoRA obtained through local training, we choose a diverse set of content images from the DreamBooth~\cite{ruiz2023dreambooth} dataset, each containing 4-5 images of a given subject. For style, we select the previous dataset provided by the authors of StyleDrop~\cite{sohn2023styledrop} and include several classic masterpieces along with some modern innovative styles. For each style, we only use a single image for training. 

\myPara{Experimental details.}
We perform our experiments using the SDXL v1.0 base model and FLUX model and test the performance of K-LoRA using locally trained LoRA and community-trained LoRA. For the community-trained LoRA, we use the widely available LoRA models from Hugging Face for testing. For the locally trained LoRA, we base on the method outlined in ZipLoRA~\cite{ZipLoRA} to obtain a set of style and content LoRAs.
For the hyperparameters mentioned in \eqref{hypepara}, we set $\alpha = 1.5$ and $\beta = 0.5$. 
This configuration was found to work effectively for nearly all cases, yielding consistently good generation results.

\subsection{Results}

\myPara{Quantitative comparisons.}
We randomly select 18 combinations of objects and styles, each of which consists of 10 images to perform quantitative comparisons. We use CLIP~\cite{clip} to measure the style similarity. We compute the subject similarity through CLIP score and DINO score~\cite{zhang2022dino}. We compare our method with popular approaches in the community as well as state-of-the-art methods, including direct arithmetic merging, joint training, ZipLoRA~\cite{ZipLoRA}, and B-LoRA~\cite{B-LoRA}. The results are shown in~\tabref{table:alignment}. It can be observed that our method significantly improves subject similarity metrics compared to previous approaches, while also achieving satisfactory style similarity. 

\begin{table}[h!]
    \centering
    \footnotesize
        \begin{tabular}{lccc}
            \toprule
            \textbf{Method} & \textbf{Style Sim $\uparrow$} & \textbf{CLIP Score $\uparrow$} & \textbf{DINO Score $\uparrow$} \\
            \midrule
            \textbf{Direct} & 48.9\% & 66.6\% & 43.0\% \\
            \textbf{Joint}  & 68.2\% & 57.5\% & 17.4\% \\
            \textbf{B-LoRA}~\cite{B-LoRA} & 58.0\% & 63.8\% & 30.6\% \\
            \textbf{ZipLoRA}~\cite{ZipLoRA} & 60.4\% & 64.4\% & 35.7\% \\
            \textbf{K-LoRA (ours)} & 58.7\% & 69.4\% & 46.9\% \\
            \bottomrule
        \end{tabular}%
    \caption{\textbf{Quantitative comparisons.} Comparison of alignment results across different methods.}
    \label{table:alignment}
\end{table}

\begin{figure*}[]
    \centering    
    \setlength{\abovecaptionskip}{2pt}
    \includegraphics[width=0.9\linewidth]{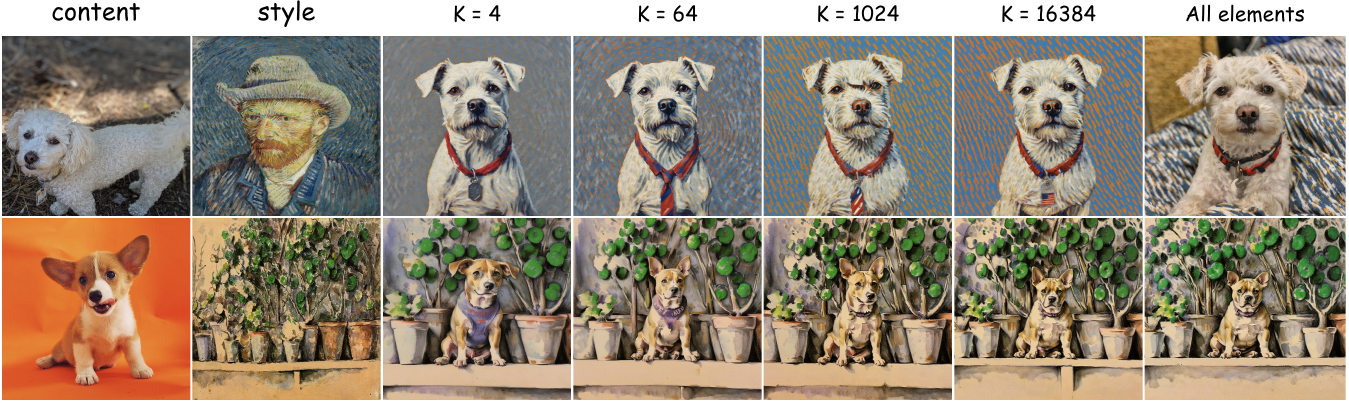}
    \caption{\textbf{Selection of $K$.}  Evaluations of the impact on different $K$ in \ourMthd{}.}
    \label{fig:K}
\end{figure*}

\myPara{Qualitative comparisons.}
In order to ensure a fair evaluation, all experiments at this stage are conducted using SD, the result is shown in \figref{fig:compare}, the method for merging LoRAs~\cite{Mergingloras} struggles to preserve the original shape, color, and stylistic features of the object when the fusion ratio is set directly to 1:2 without extensive parameter adjustments or seed selection. B-LoRA~\cite{B-LoRA} mainly captures the color and appearance of objects in the original image, often leading to overfitting of the color, which makes it difficult to distinguish the original objects in the generated image.
In ZipLoRA~\cite{ZipLoRA} and joint training methods, while certain stylistic textures are incorporated, the model tends to focus on the background elements of the style rather than capturing the style itself, resulting in a lower success rate.
In contrast, our method addresses these limitations by producing higher-quality output images with stable performance across a wide range of seed variations. Additionally, our approach eliminates the need for extra training or parameter fine-tuning.

We present a randomly selected set of 22 results to users for comparative evaluation. Each set includes outputs from ZipLoRA, B-LoRA, and our method, along with reference images for both the training subject and the style. Users were asked to identify which method best preserves both style and subject. The results, shown in \tabref{tab:user_study}, indicate that our method is the most preferred. In addition, we consulted with GPT-4o for a similar assessment. Our method shows a significant advantage in GPT-4o evaluations, further reflecting the superiority of our method.

\begin{table}[ht]
\centering
\small
\begin{tabular}{lcc}
\toprule
\textbf{Method} & \textbf{User Preference} & \textbf{GPT-4o Feedback} \\
\midrule
ZipLoRA~\cite{ZipLoRA}          &       29.2\%      &       5.6\%     \\
B-LoRA~\cite{B-LoRA}           &       18.1\%        &     11.1\%        \\
Ours             &      52.7\%          &     83.3\%        \\
\bottomrule 
\end{tabular}
\caption{\textbf{User study results and GPT-4o feedback.} }
\label{tab:user_study}
\end{table}

\begin{figure}[]
    \centering
    \setlength{\abovecaptionskip}{2pt}
    \includegraphics[width=\linewidth]{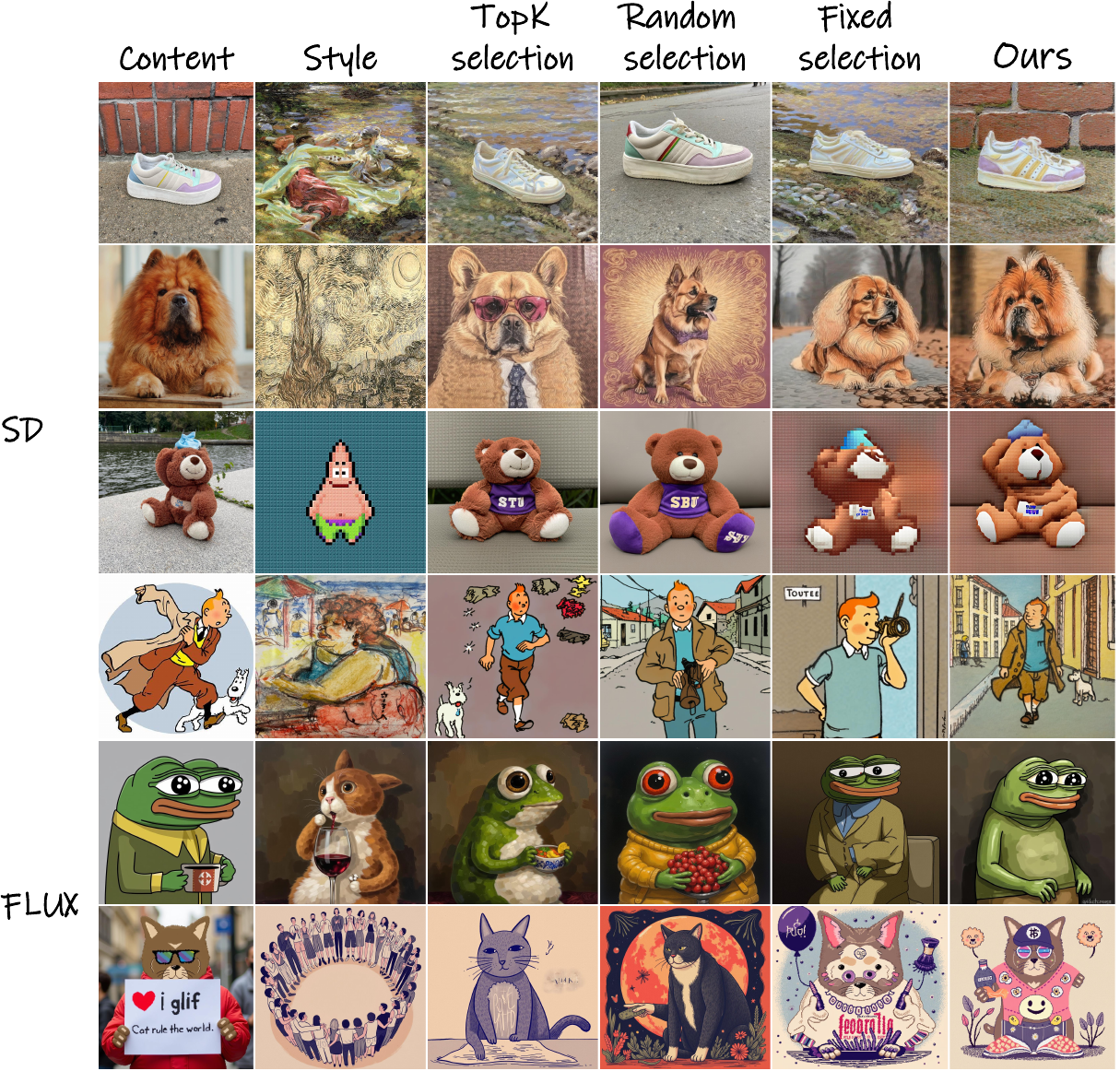}
    \caption{\textbf{Ablation of Top-K selection and scaling factor.} We compare different methods using five sets of images. The four rows above represent the results of SD, while the rows below present the results of FLUX, which include both locally trained LoRA and community trained LoRA.
}
    \label{fig:ablation}
        \vspace{-5pt}
\end{figure}

\subsection{Ablation Analysis}

\myPara{Top-K selection.} We conduct two experiments to validate the effectiveness of the Top-K selection method: fixed selection and random selection.
Finding (ii) suggests a straightforward approach: If the scale factor is greater than one, the content LoRA is selected; Otherwise, the style LoRA is chosen. This approach, which we refer to as ``Fixed Selection'' serves as a useful baseline to test the ablation of the Top-K selection method.
It can also be seen as an extension and refinement of Multi-LoRA composition~\cite{Multi-LoRAComposition}, which has shown promising results in certain scenarios. 
However, under specific style LoRA conditions, this method may result in object blurring or alterations in the content’s appearance, as shown in \figref{fig:ablation}.

To ensure that our module performs consistently within the specified forward layer arrangement rather than relying on arbitrary configurations, we conduct a controlled experiment termed ``Random Selection'' using a random seed. In this setup, the model uses a random number with a 1/3 probability of selecting content attention and a 2/3 probability of selecting style attention. As shown in \figref{fig:ablation}, under these random selection conditions, the generated images often retain only a single object feature or style feature, or fail to maintain either altogether. This outcome further validates our finding (ii), highlighting the distinct roles played by object and style components at earlier and later diffusion time steps, respectively.

Furthermore, we evaluate the impact of different choices of $K$ on the generated images, as illustrated in \figref{fig:K}. Within the Top-K approach, we systematically vary the values of $K$. Our observations indicate that when $K$ is relatively small, neither the style nor the characteristics of the object are sufficiently prominent. This issue gradually improves as $K$ increases. However, if $K$ becomes excessively large, the style may not be preserved, and the shape of the object can undergo significant distortions.

\myPara{Scaling factor.} To evaluate the effectiveness of the scaling factor, we remove it and focus solely on the original Top-K approach. In the first experiment, as shown in \figref{fig:ablation}, our analysis reveals that while the exclusive use of Top-K can produce satisfactory results under certain conditions, expanding the experimental scope uncovers instances of object distortion and style loss.
To further assess the significance of gamma within the scaling factor, we test the performance of two LoRA models with distinct sources, characterized by substantial differences in their element sums. As illustrated in the bottom row of \figref{fig:ablation}, it is evident that Top-K selection  fails to capture the style accurately, while the fusion of object and style in fixed selection is noticeably weaker compared to our approach. We also experiment with an alternative scale. The detailed procedure is provided in the supplementary material (Sec.~\ref{new-scale}).

In conclusion, the removal of these two modules leads to a decrease in generative performance, underscoring their critical contributions to the overall effectiveness of the model. 

\section{Conclusions}
In this paper, we introduce \ourMthd{}, which can seamlessly merge independently trained style and subject LoRA models.
\ourMthd{} enables precise object fine-tuning while preserving the intricate details of the original style.
Our approach effectively leverages the contributions of both object and style LoRAs at each diffusion step through Top-K selection and scaling factors, maximizing the use of the original weights and
allowing for accurate style fusion without the need for retraining or manual hyperparameter tuning.
%

{
    \small
    \bibliographystyle{ieeenat_fullname}
    \bibliography{main}
}

\newpage
\appendix

\begin{center}
     \Large\textbf{Supplementary Material}
\end{center}

\noindent The supplementary material is structured as follows:

\begin{enumerate}
    \item We first evaluated our results on extensive datasets and community LoRAs on different models to validate the effectiveness of our approach in section \ref{visual-result}.
    \item We compared our method with the other methods in section \ref{compare}.
    \item We assessed the influence of complex prompts on the model's performance in section \ref{prompt-control}.
    \item We experimented with a new scale and tested its comparative effects in section \ref{new-scale}.
    \item We utilized Community LoRA in combination with local LoRA to conduct integrated performance evaluations and examined random seeds on model performance through comprehensive testing in section \ref{robust}.
    \item We tested the choice of different parameters in scale factors in section \ref{addition-ablation}.
    
\end{enumerate}

\section{Visual Results}\label{visual-result}

\begin{figure*}
    \centering
    \includegraphics[width=1\linewidth]{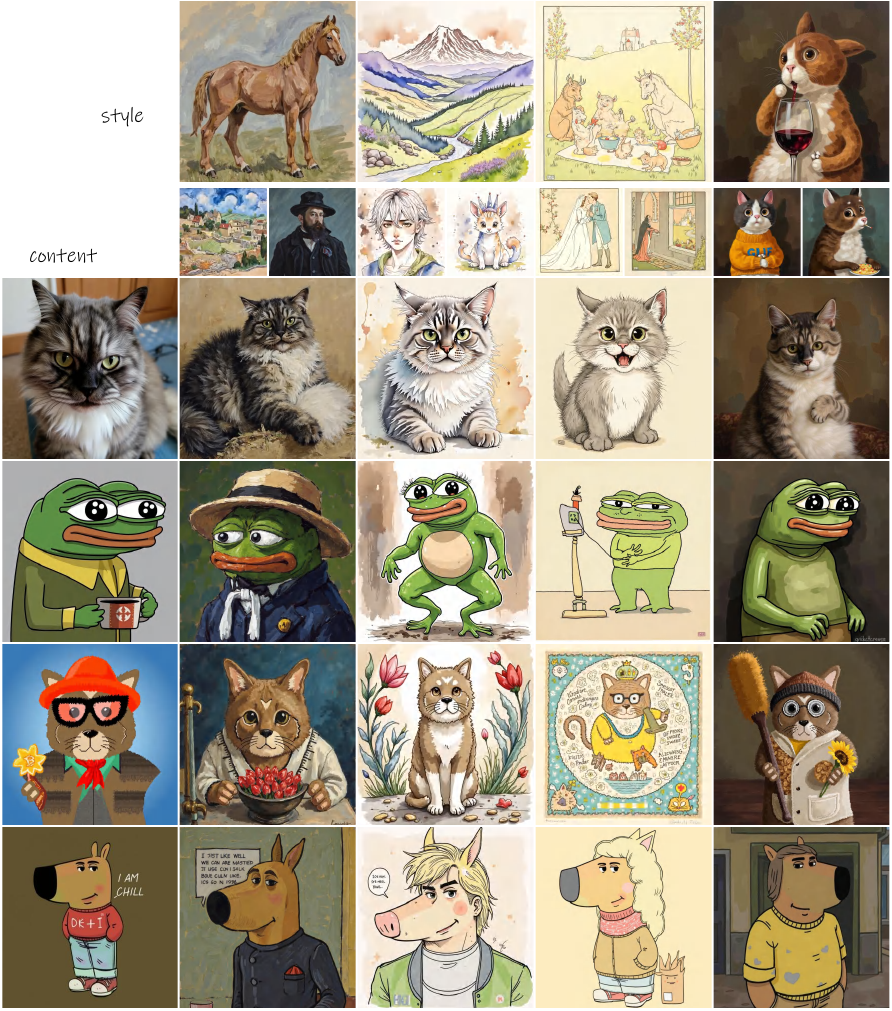}
    \caption{
    \textbf{Additional Generated Results using FLUX.} The images in each position correspond to the object above and the style on the left, showing the results generated by applying the different LoRAs with our method.
    }
    \label{picturemap3}
\end{figure*}

\begin{figure*}
    \centering
    \includegraphics[width=1\linewidth]{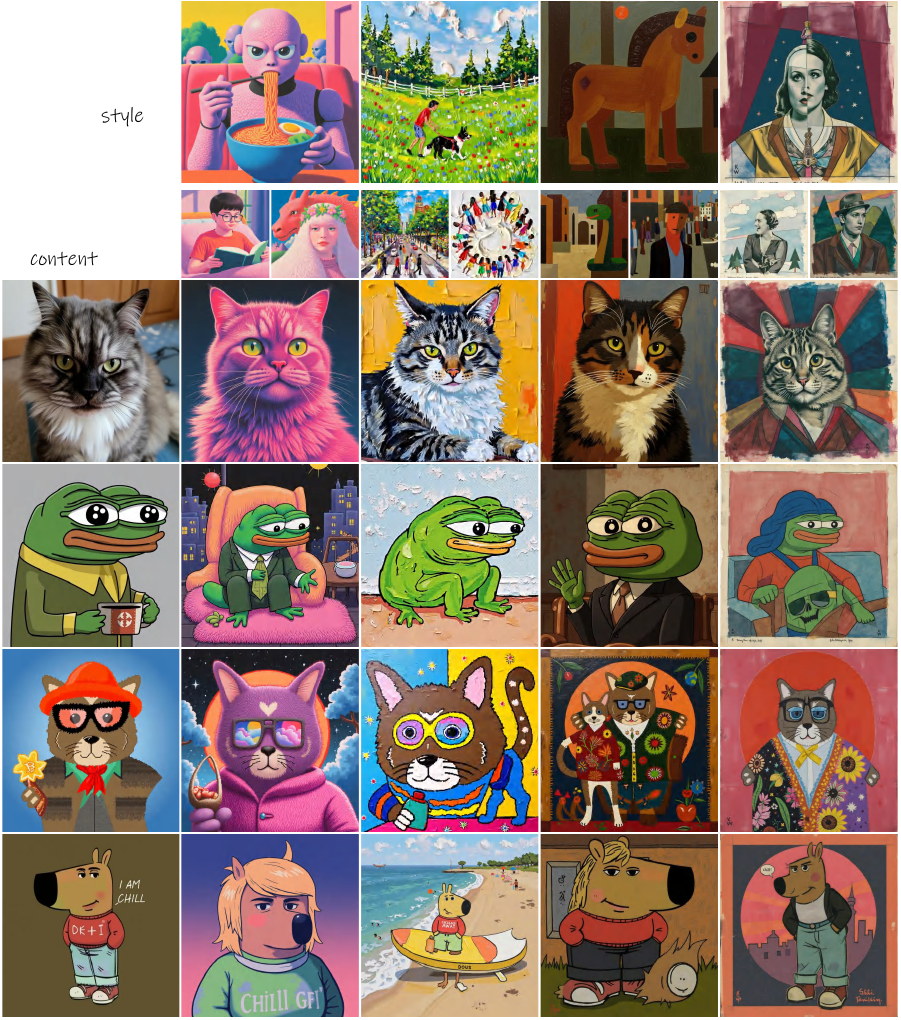}
    \caption{\textbf{Additional Generated Results using FLUX.} The images in each position correspond to the object above and the style on the left, showing the results generated by applying the different LoRAs with our method.}
    \label{picturemap4}
\end{figure*}

\begin{figure*}
    \centering
    \includegraphics[width=1\linewidth]{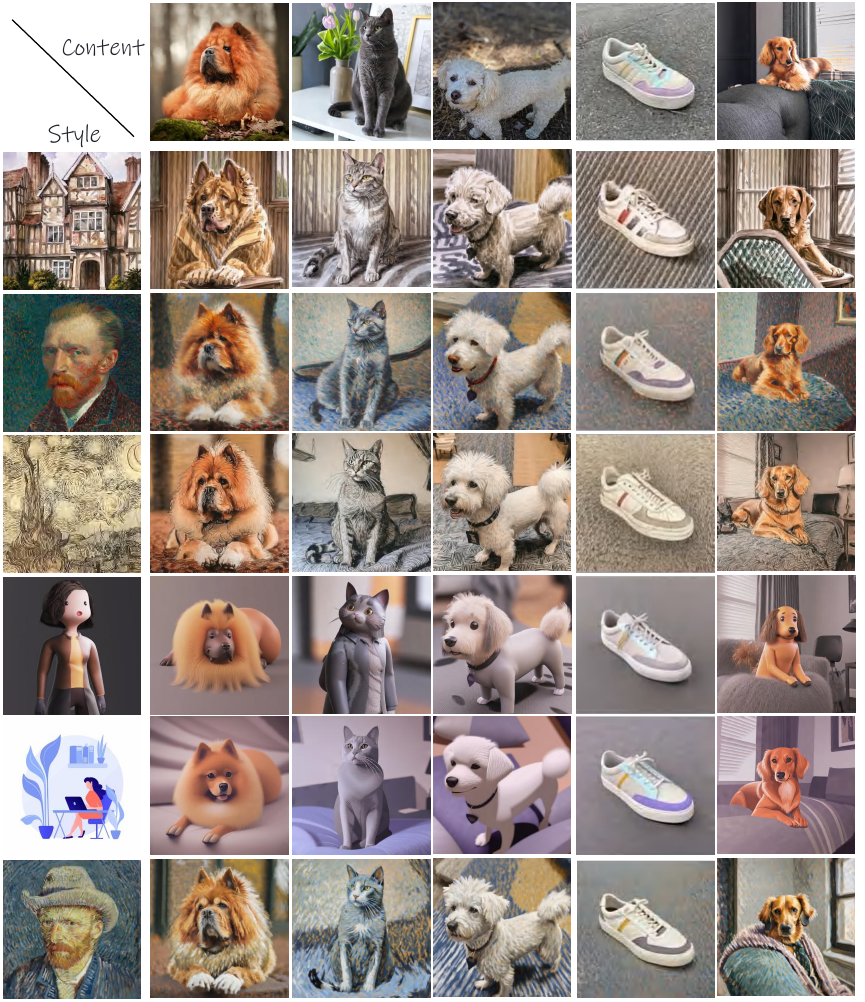}
    \caption{
    \textbf{Additional Generated Results using SD.} The images in each position correspond to the object above and the style on the left, showing the results generated by applying the different LoRAs with our method.
    }
    \label{picturemap1}
\end{figure*}

\begin{figure*}
    \centering
    \includegraphics[width=1\linewidth]{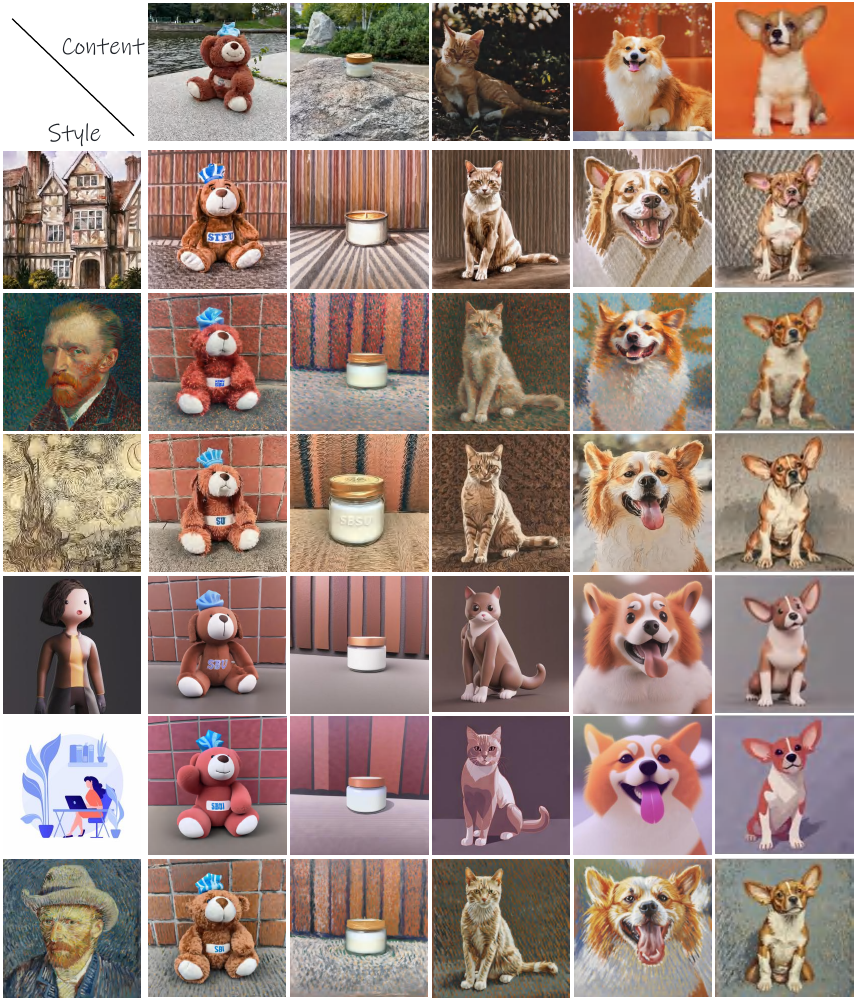}
    \caption{\textbf{Additional Generated Results using SD.} The images in each position correspond to the object above and the style on the left, showing the results generated by applying the different LoRAs with our method.}
    \label{picturemap2}
\end{figure*}

We employ datasets from StyleDrop~\cite{sohn2023styledrop} and DreamBooth~\cite{ruiz2023dreambooth} with Stable Diffusion (SD), as depicted in \figref{picturemap1} and \figref{picturemap2}, we also evaluated our method on FLUX using LoRAs from Hugging Face, as shown in \figref{picturemap3} and \figref{picturemap4}.
By systematically combining these object and style LoRAs, we obtained a sequence of images that demonstrates the effectiveness of our approach in seamlessly integrating both object and style, yielding consistent and high-quality visual outputs.

\section{Additional Comparisons}\label{compare}

\begin{figure*}
    \centering
    \includegraphics[width=1\linewidth]{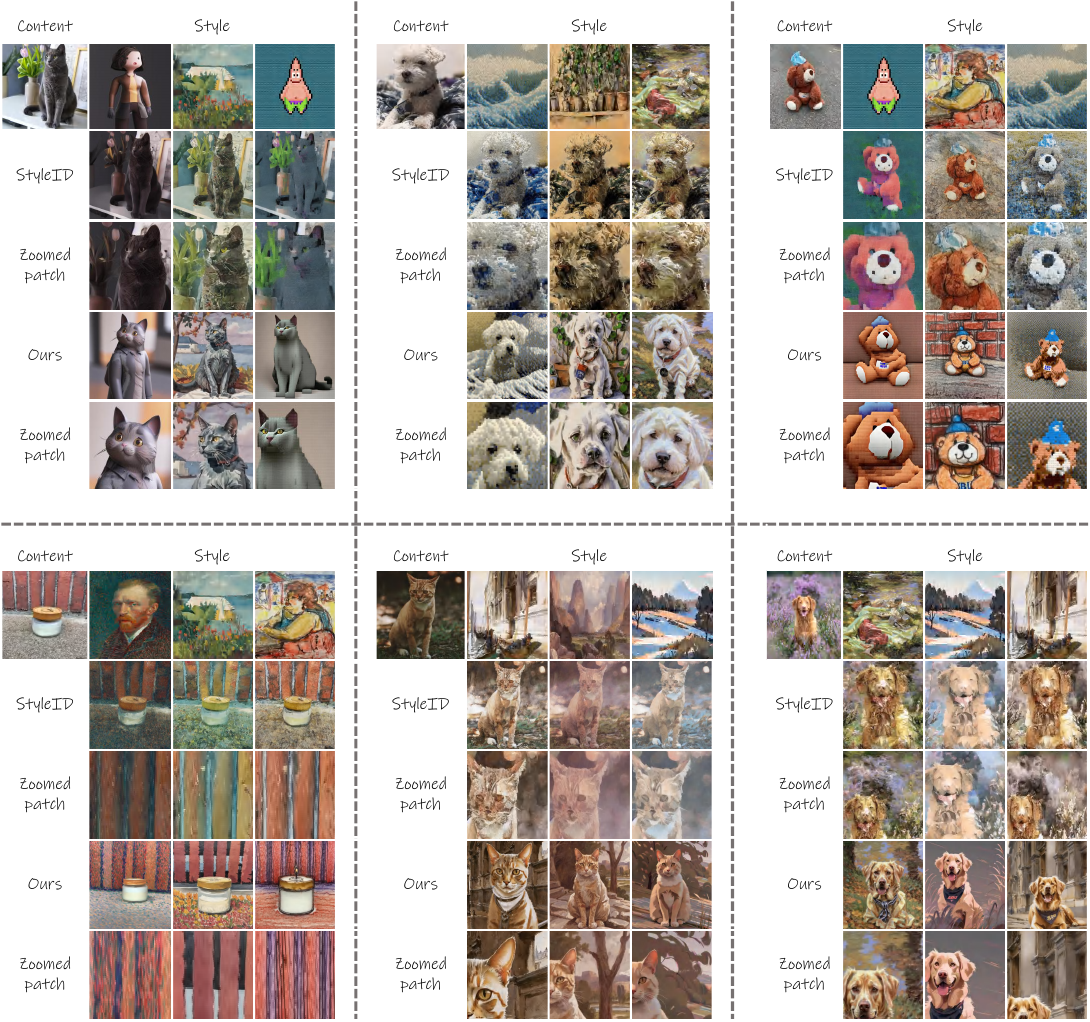}
    \caption{
    \textbf{Additional Comparisons.} We compare the StyleID~\cite{2023arXiv231209008C} method and then capture zoomed patches in the output image to observe detailed texture information and stylistic features. Within each block, the second and third rows represent StyleID results along with its corresponding zoomed patch, while the subsequent two rows illustrate the result of our method and the associated zoomed patch.
    }
    \label{details}
\end{figure*}

We have added a comparison with StyleID~\cite{2023arXiv231209008C}, as shown in \figref{details}. It can be observed that StyleID~\cite{2023arXiv231209008C} effectively achieves style transfer while preserving texture quality. However, the generated objects might be slightly blurred or the style generated may not be distinct. Additionally, compared to our method, their approach is based on the fixed layout of original image, which may not generalize well to backgrounds and actions.

\section{Prompt Control}\label{prompt-control}
\begin{figure*}
    \centering
    \includegraphics[width=1\linewidth]{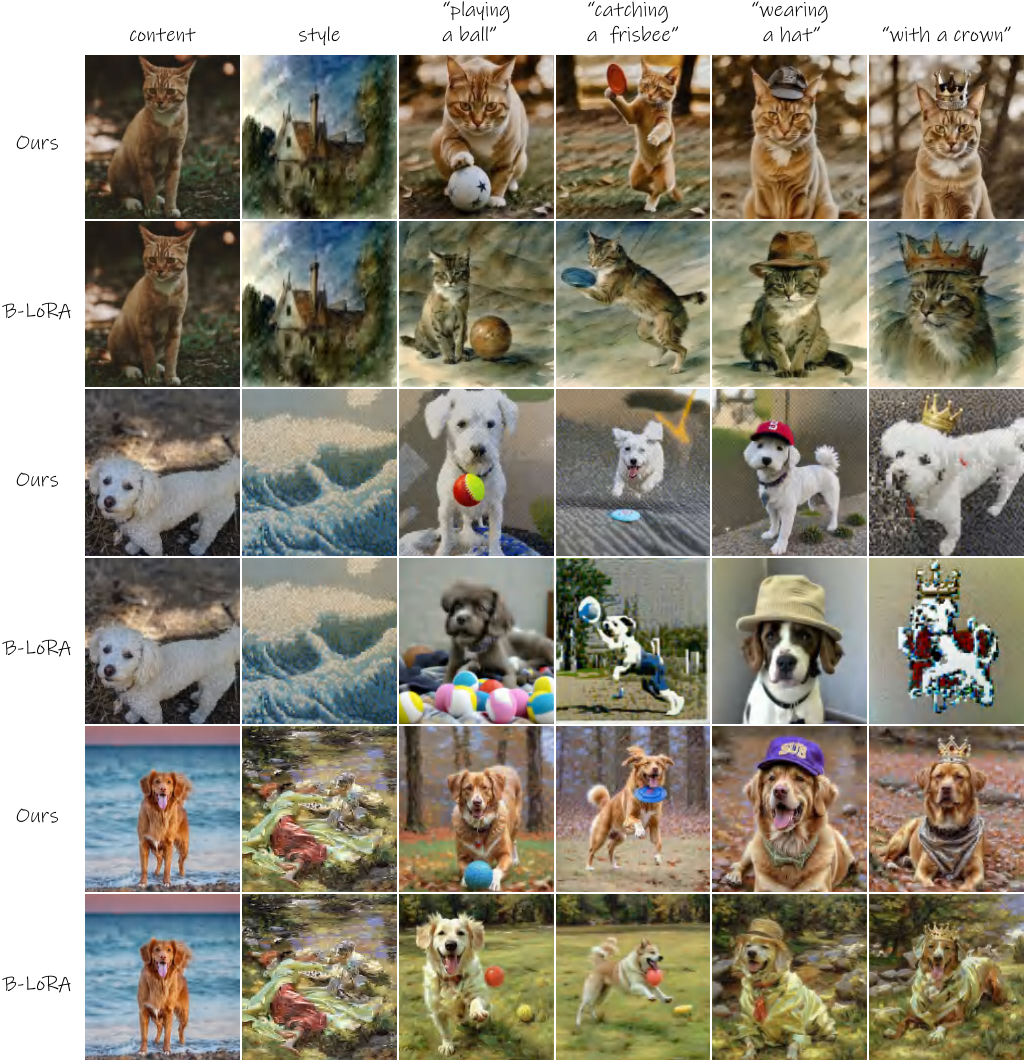}
    \caption{
    \textbf{Prompt Control.} We introduce prompts for new scenes, new actions, and new objects to validate our method's ability to re-contextualize content and maintain stylistic consistency.
    }
    \label{prompt1}
\end{figure*}

\begin{figure*}
    \centering
    \includegraphics[width=1\linewidth]{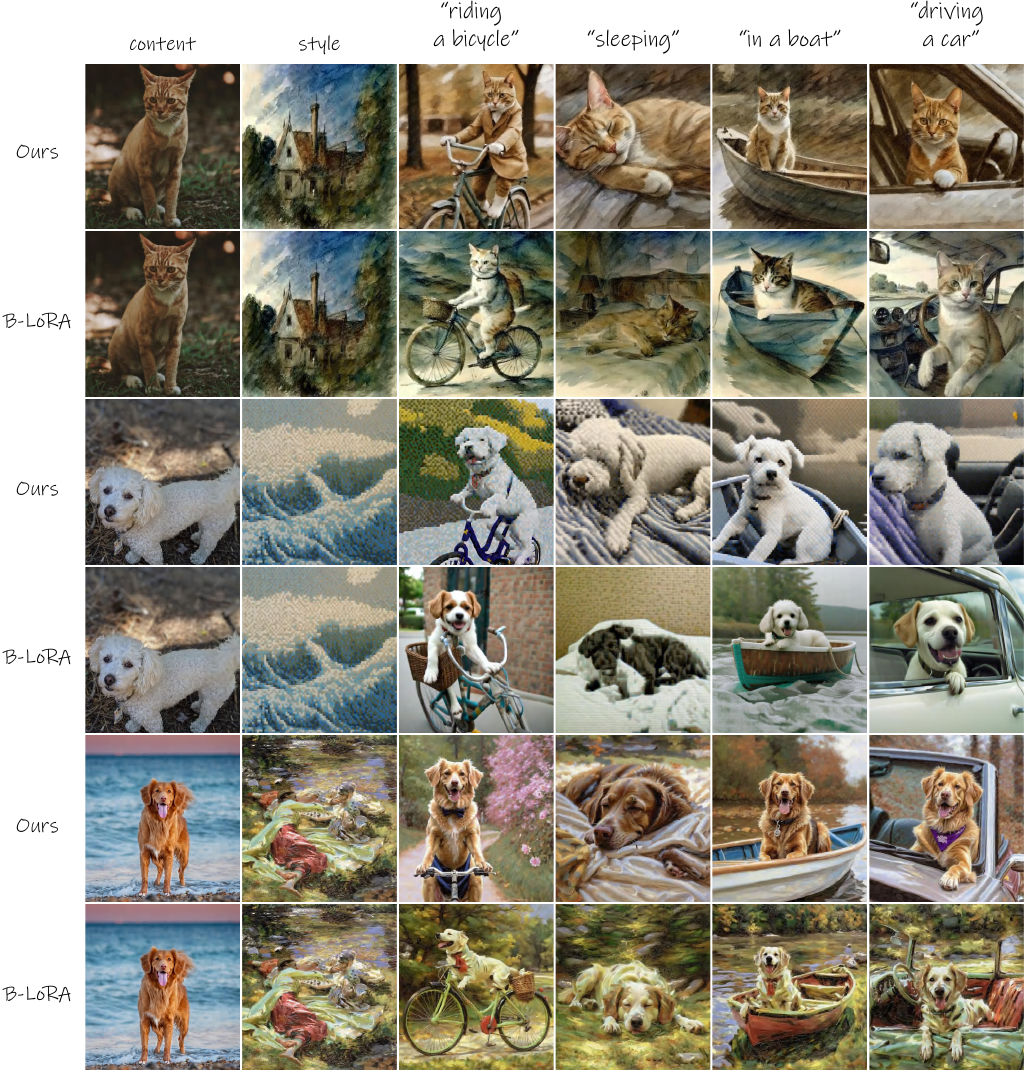}
    \caption{
    \textbf{Prompt Control.} We introduce prompts for new scenes, new actions, and new objects to validate our method's ability to re-contextualize content and maintain stylistic consistency.
    }
    \label{prompt2}
\end{figure*}

We conduct experiments to evaluate whether our method can modify the object's actions, the surrounding environment, or introduce new elements through prompt adjustments. As illustrated in \figref{prompt1} and \figref{prompt2}, after modifying the prompts, our method effectively retains the original object's features and stylistic attributes, while also integrating new elements or scene details seamlessly.

\section{New Scale} \label{new-scale}

In the main text of our paper, we employ the scale given by \eqref{hypepara} as follows:
\begin{equation}
    S = \alpha  \cdot \frac{{t}_{{now}}}{{t}_{{all}}} + \beta.
\end{equation}
Inspired by \cite{ProSpect}, we also introduce an alternative scale factor:
\begin{equation}
    S^{*} = \left( \alpha'  \cdot \frac{{t}_{{now}}}{{t}_{{all}}} + \beta' \right) \%  \ \alpha.
    \label{hypepara1}
\end{equation}
In this equation, we set $\alpha' = 1.5$ and $\beta' = 1.3$, which means that the style information is enhanced to some extent at the beginning of the generation process, allowing the model to capture certain block information from the style LoRA. \figref{fig:scale} below illustrates the primary differences between the two scales. 

For $S^{*}$ results, since the style information is enhanced during the early diffusion steps, the generated images capture the background and color block information from the style LoRA. However, this approach results in a weakened learning effect for the texture and brushstrokes information in the style LoRA. This represents a trade-off, and users can select different scale factors based on their preferences.
\begin{figure}[h]
    \includegraphics[width=\linewidth]{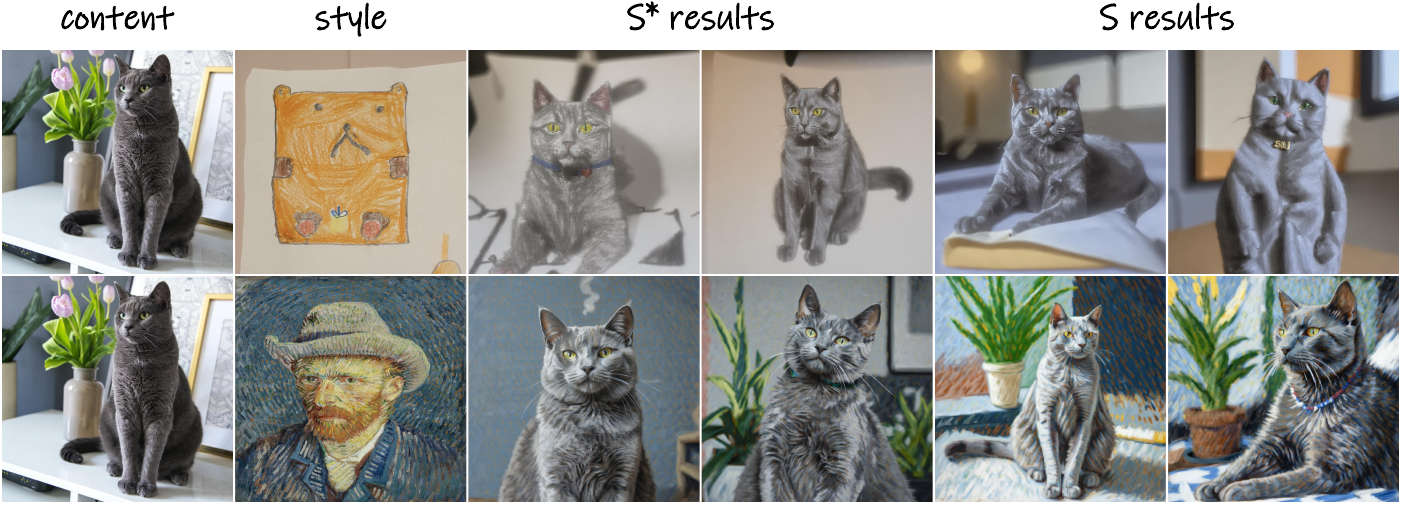}
    \caption{\textbf{Results of different scaling factors.} Corresponding generation results of \ourMthd{} with differernt scaling factor and for each object-style pair, two seeds are randomly selected.
}
    \label{fig:scale}
\end{figure}

\section{Robustness Analysis}\label{robust}

\begin{figure*}
    \centering
    \includegraphics[width=1\linewidth]{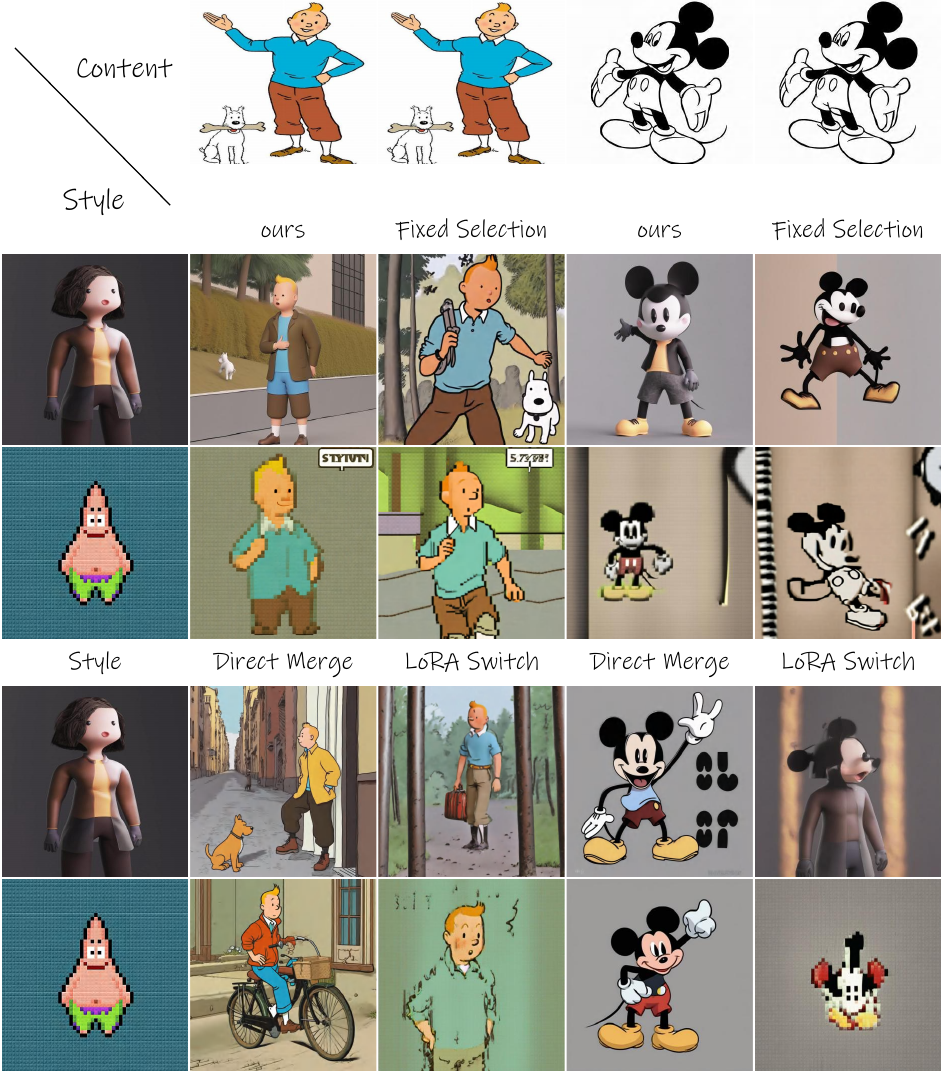}
    \caption{
\textbf{Robustness Validation.} We utilize community LoRAs and locally trained LoRAs to compare the Fixed Selection proposed in the main text, direct arithmetic merging LoRA as a baseline comparison, Multi-LoRA Composition~\cite{Multi-LoRAComposition} methods, in order to validate generalizability and robustness.
    }
    \label{community}
\end{figure*}

\begin{figure*}
    \centering
    \includegraphics[width=1\linewidth]{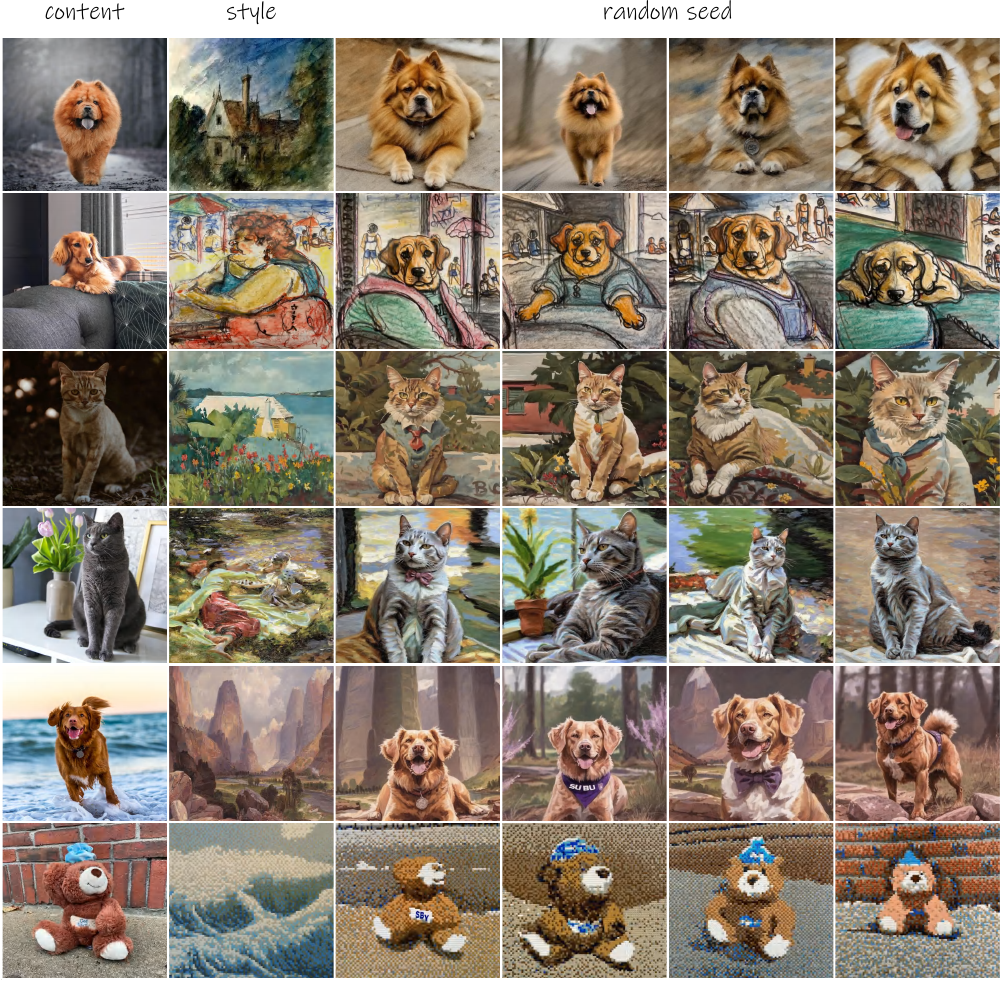}
    \caption{
        \textbf{Robustness Validation.} We randomly select seeds to further validate stability.
    }
    \label{randomseed}
\end{figure*}

We evaluate LoRA models from various sources, where the object LoRA is sourced from the community, while the style LoRA is trained locally. We also compare DirectMerge~\cite{Mergingloras}, Multi-LoRA composition~\cite{Multi-LoRAComposition}, and our proposed Fixed Selection approach. As shown in \figref{community}, our method demonstrates superior performance in learning both object and style characteristics, surpassing other approaches. Furthermore, we test the robustness of our approach by selecting random seeds to assess stability. The results, presented in \figref{randomseed}, indicate that our method consistently achieves stable fusion across a broad range of seed selections, ensuring reliable integration.

\section{Additional Ablations}\label{addition-ablation}

In the main text, we employe a scale with two hyperparameters, $\alpha$ and $\beta$. Specifically, we set $\alpha$ to 1.5 and $\beta$ to 0.5, enabling objects and styles to exert varying levels of influence at different positions. To validate the suitability of the selected parameters, we compute the CLIP similarity scores between 18 randomly chosen sets of generated images and their corresponding original object/style references. 
The results shown in the table below represent the summation of CLIP similarity scores.

\begin{table}[h]
\centering
\footnotesize
\label{tab:alpha_beta}
\begin{tabular}{c|ccc}
    $\beta \backslash \alpha$ & 1.0    & 1.5     & 2.0    \\
    \hline
    0.25      & 125.3\%  & 126.7\%  & 127.0\% \\
    0.50      & 126.5\% &\textbf{ 128.1\% }  & 126.2\% \\
    0.75      & 124.5\% & 125.8\% & 125.3\%  \\

\end{tabular}  
\end{table}

We can see that the optimal setting for $\alpha$ and $\beta$ is 1.5 and 0.5, respectively. This weight configuration satisfies almost all content-style pairs according to our experiments, and users do not need to make further adjustments. 

\end{document}